%% file: main.tex
\documentclass[10pt,twocolumn,letterpaper]{article}

\usepackage[pagenumbers]{cvpr} 


\definecolor{cvprblue}{rgb}{0.21,0.49,0.74}
\usepackage[pagebackref,breaklinks,colorlinks,allcolors=cvprblue]{hyperref}

\usepackage{listings}
\usepackage{epsfig}
\usepackage{amsmath}
\usepackage{amssymb}
\usepackage{verbatim}
\usepackage{booktabs}
\usepackage{tabularx}
\usepackage{amsfonts}
\usepackage{multirow}
\usepackage{bbding}
\usepackage{algorithm,algorithmicx,algpseudocode}
\usepackage{array}
\usepackage{colortbl}
\usepackage{natbib}
\usepackage{wasysym}
\usepackage{wrapfig}
\usepackage{pifont}
\newcommand{\cmark}{\ding{51}}%

\definecolor{mygray}{gray}{.92}
\definecolor{lightgray}{gray}{.96}
\definecolor{myy}{RGB}{126,95,0}
\definecolor{ggray}{RGB}{127,127,127}
\definecolor{mygreen}{RGB}{0,0,0}
\definecolor{myred}{RGB}{240,16,89}
\definecolor{myblue}{RGB}{0,114,188}
\definecolor{darkgreen}{rgb}{0.0, 0.5, 0.0}
\definecolor{demphcolor}{RGB}{100,100,100}

\title{AllRestorer: All-in-One Transformer for Image Restoration under Composite Degradations}

\author{%
\bf \textbf{Jiawei Mao}$^{1}$
\quad
\textbf{Yu Yang}$^{1}$ \quad
\textbf{Xuesong Yin}$^{1}$ \quad 
\textbf{Ling Shao}$^2$ \\
\bf \textbf{Hao Tang}$^3$ \quad \vspace{1.em} \\
$^1$Hangzhou Dianzi University
\quad $^2$University of the Chinese Academy of Sciences \quad $^3$Peking University
}

\begin{document}
\twocolumn[{%
\renewcommand\twocolumn[1][]{#1}%
\maketitle

\begin{center}
\centering
\vspace{-0.4cm}
\setlength{\belowcaptionskip}{6mm}
\includegraphics[width=1\linewidth]{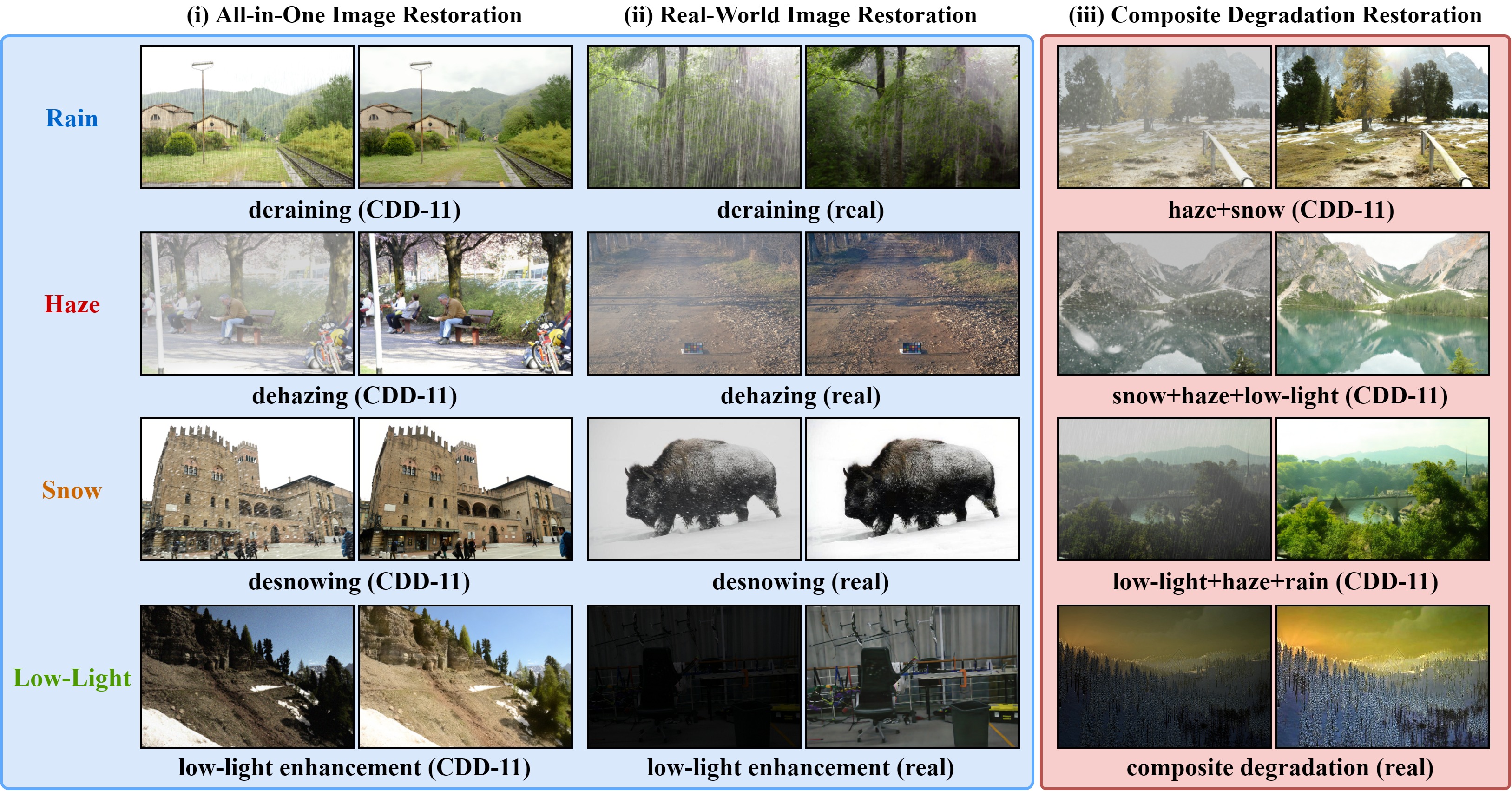}
\captionof{figure}{
\textbf{AllRestorer} can be applied to several image restoration tasks. \textbf{(i) All-in-One Image Restoration:} AllRestorer can address multiple single degradation scenarios with just one set of parameters.  \textbf{(ii) Real-World Image Restoration:} AllRestorer can successfully respond to real-world all-in-one restoration and composite degradation restoration challenges. 
\textbf{(iii) Composite Degradation Restoration:} AllRestorer can be applied to various composite degradation restoration tasks with only a single network.}
\label{fig1}
\end{center}
}]

\input{sec/0_abstract}  
\input{sec/1_intro}
\input{sec/2_related_work}
\input{sec/3_method}
\input{sec/4_experiment}

\input{sec/5_conclusion}

\clearpage
{
    \small
    \bibliographystyle{ieeenat_fullname}
    \bibliography{main}
}


\end{document}

%% file: sec/0_abstract.tex
\begin{abstract}

Image restoration models often face the simultaneous interaction of multiple degradations in real-world scenarios. Existing approaches typically handle single or composite degradations based on scene descriptors derived from text or image embeddings. However, due to the varying proportions of different degradations within an image, these scene descriptors may not accurately differentiate between degradations, leading to suboptimal restoration in practical applications.
To address this issue, we propose a novel Transformer-based restoration framework, \textbf{AllRestorer}. In AllRestorer, we enable the model to adaptively consider all image impairments, thereby avoiding errors from scene descriptor misdirection. Specifically, we introduce an All-in-One Transformer Block (AiOTB), which adaptively removes all degradations present in a given image by modeling the relationships between all degradations and the image embedding in latent space.
To accurately address different variations potentially present within the same type of degradation and minimize ambiguity, AiOTB utilizes a composite scene descriptor consisting of both image and text embeddings to define the degradation. Furthermore, AiOTB includes an adaptive weight for each degradation, allowing for precise control of the restoration intensity.
By leveraging AiOTB, AllRestorer avoids misdirection caused by inaccurate scene descriptors, achieving a \textbf{5.00 dB} increase in PSNR compared to the baseline on the CDD-11 dataset.
\end{abstract}

%% file: sec/1_intro.tex
\section{Introduction}
\label{sec:intro}

To meet the demands of autonomous driving~\cite{fu2024drive,zhou2024drivinggaussian,xu2024drivegpt4} under various road conditions, recent studies have moved beyond image restoration for single-degradation scenarios~\cite{Zamir2021Restormer,Mou2022DGUNet,song2023srudc} and have shifted their focus to all-in-one image restoration tasks~\cite{AirNet,potlapalli2023promptir,zhu2023Weather}. As an emerging low-level vision task, all-in-one restoration requires the model to restore visuals from various types of impaired inputs.

While all-in-one image restoration has shown satisfactory performance under experimental conditions, real-world scenarios often involve multiple degradations coexisting in the same scene, posing a significant challenge to most restoration methods. To address this issue, there is an increasing need for a solution that combines all-in-one image restoration with composite degradation restoration.
Recent approaches introduce the concept of scene descriptors in all-in-one image restoration~\cite{guo2024onerestore,chen2024teaching}, which guide the model to accurately perform restoration by providing finer distinctions between single and composite degradations.

VPIP~\cite{chen2024learning} introduces paired degraded and clean images corresponding to the current restoration task as visual scene descriptors, providing prior information for restoring composite degraded images. However, the VPIP setting requires users to accurately determine the type of degradation and provide the corresponding visual scene descriptors, making it impractical for real-world deployment.
To address this limitation, MPerceiver~\cite{ai2024multimodal} introduces impaired image-to-text inversion, creating a textual scene descriptor that adaptively leverages diffusion-based prior knowledge for composite image restoration without human intervention. OneRestore~\cite{guo2024onerestore} uses explicit definitions of different degradations in textual form to guide scene descriptors, enabling accurate restoration of composite degradations.

Despite some progress under experimental conditions, these schemes do not effectively model composite degradations. Since composite degradations include multiple degradations, and the proportions of these degradations vary, it often leads to confusion for users and models in identifying the correct degradation type, causing incorrect restoration.

 \begin{figure}[t]
    \centering
    \includegraphics[width=1\linewidth]{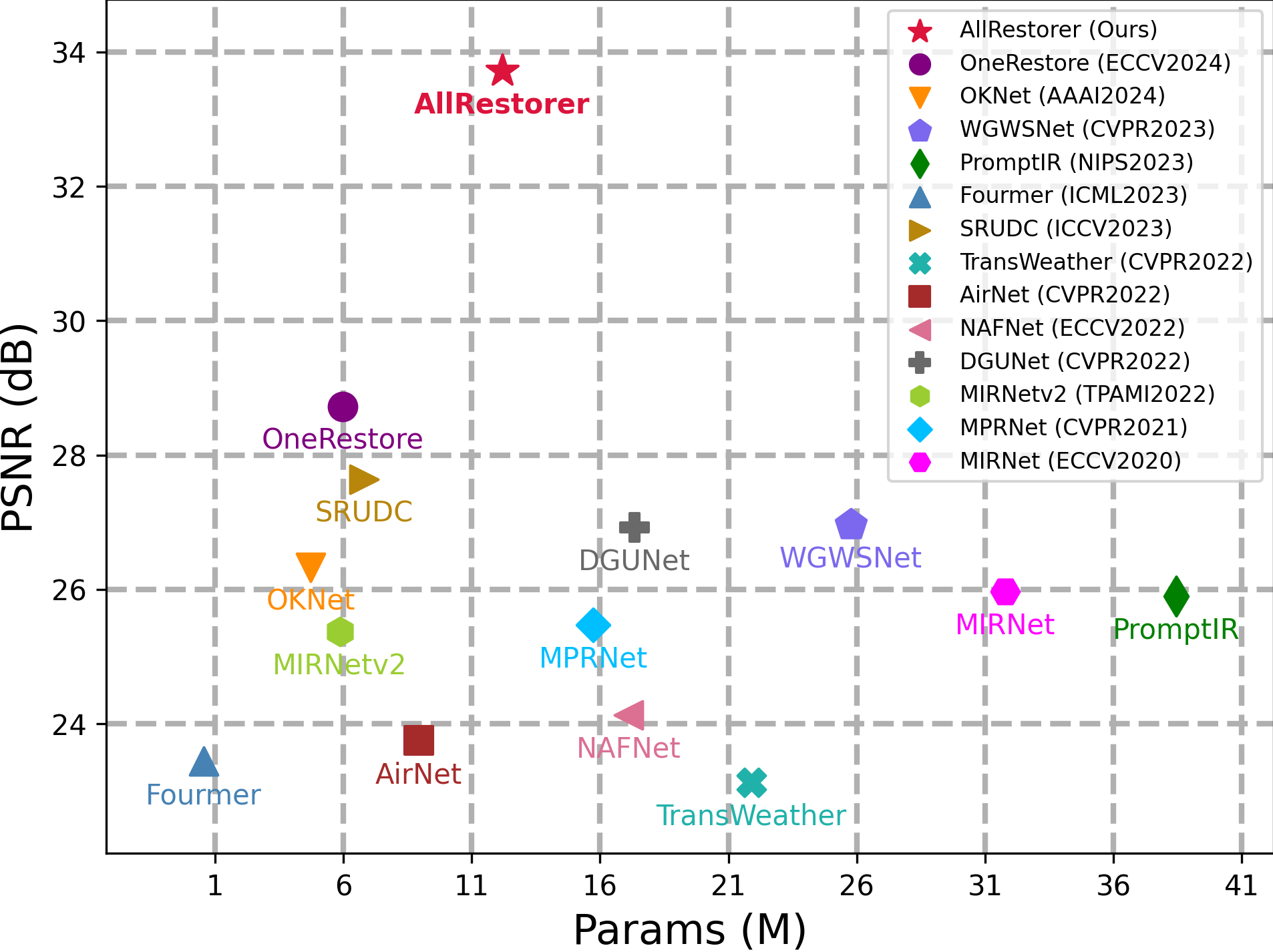}
     \caption{
     AllRestorer achieves state-of-the-art performance on the CDD-11 dataset while remaining lightweight.}
     \vspace{-0.2cm}
 \label{fig2}
  \end{figure}

In response to the above issues, we propose a transformer-based restoration model, \textit{AllRestorer}, designed to handle the four physical degradation types (see \cref{fig1}) in the official composite degradation dataset CDD-11~\cite{guo2024onerestore}.
In AllRestorer, we propose a novel \textit{All-in-One Transformer Block (AiOTB)} to adaptively remove all degradations in latent space, effectively addressing the composite degradation problem.
As the core component of AllRestorer, AiOTB builds a memory bank containing descriptors of all single degradation types (e.g., rain, snow, fog, and low-light) and implements a novel \textit{All-in-One Attention (AiOA)} mechanism. AiOA leverages the memory bank's descriptors to introduce restoration schemes for all degradations. The self-attention mechanism in AiOTB adaptively assigns the appropriate restoration schemes for each degradation in latent space.
Considering that visual scene descriptors lack precise definitions for degradations and that text scene descriptors may not efficiently capture the variations within the same degradation type, AllRestorer employs a composite scene descriptor that integrates both text and image embeddings.
To adapt to varying proportions of single degradations within composite degradation scenes and to support all-in-one restoration composed of multiple single-degradation processes, AiOA also includes an adaptive weighting mechanism to control the restoration intensity for different degradations.

To demonstrate the effectiveness of AllRestorer for all-in-one image restoration and composite degradation restoration, we trained and tested it on the CDD-11 dataset. As shown in \cref{fig2}, AllRestorer achieves state-of-the-art (SoTA) performance on CDD-11, surpassing the previous SoTA baseline by 5.00 dB in PSNR. Additionally, real-world tests confirmed the potential of AllRestorer for practical applications.

Our contributions can be summarized as:
\begin{itemize}
\item We propose a novel restoration scheme, AllRestorer, which effectively addresses the composite degradation challenge by removing all types of degradation.
\item Our composite scene descriptor, based on text and image embeddings, ensures an accurate representation of degradation types.
\item Our adaptive weights effectively control the restoration intensity for different degradations within composite scenes, accommodating varying proportions.
\end{itemize}

%% file: sec/2_related_work.tex
\section{Related Work}

\subsection{All-in-One Image Restoration}

 \begin{figure}[t]
    \centering
    \includegraphics[width=1\linewidth]{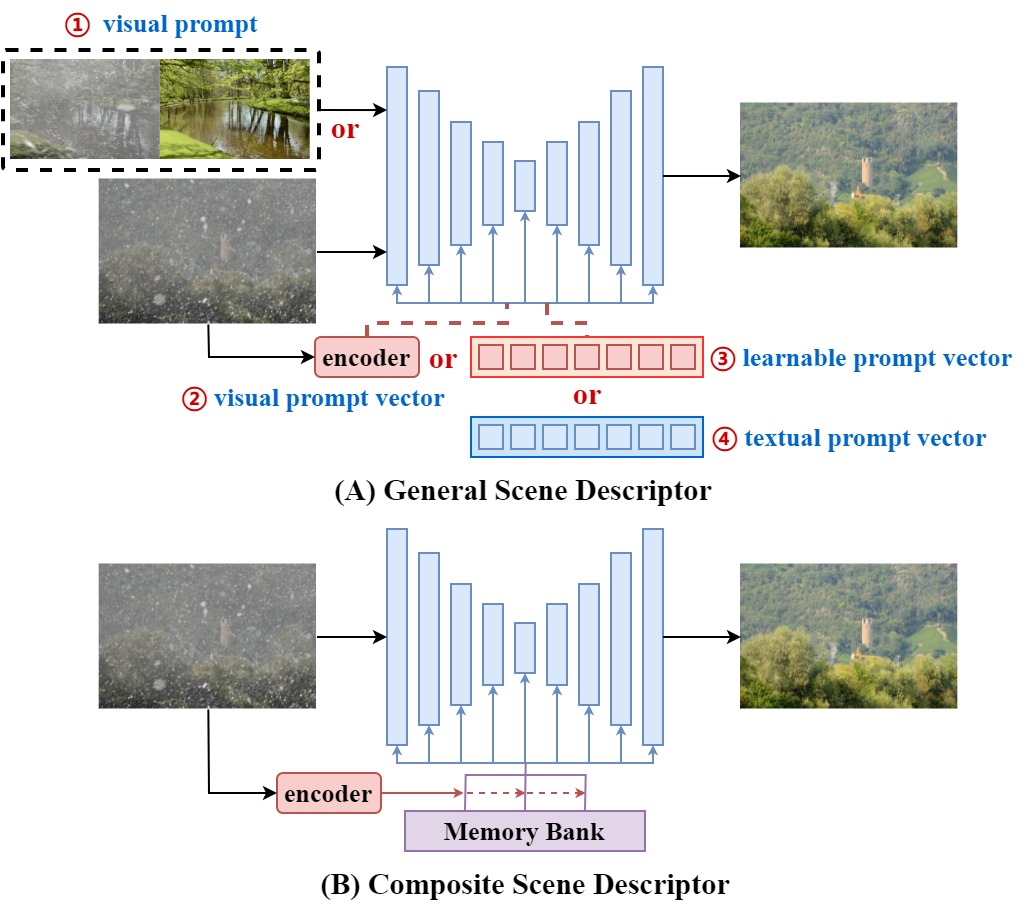}
     \caption{Scene descriptor comparison between general all-in-one restoration methods and ours.}
 \label{fig:paradigm}
      \vspace{-0.4cm}
  \end{figure}

All-in-one image restoration aims to address multiple types of degradation in real-world scenarios using a single model. All-in-One~\cite{li2020all} first introduces Neural Architecture Search (NAS) to extend the restoration task to an all-in-one approach. However, the large number of parameters in NAS makes practical deployment challenging.
TKL~\cite{chen2022learning} utilizes knowledge distillation to aggregate multiple restoration priors for the all-in-one task, while TransWeather~\cite{valanarasu2022transweather} introduces learnable queries to model different degradations. However, learnable queries lack interpretability, and knowledge distillation requires collecting teacher models.
WeatherDiff~\cite{ozdenizci2023restoring} and MPerceiver~\cite{ai2024multimodal} use diffusion models to handle all-in-one restoration. VPIP~\cite{chen2024learning} and PromptIR~\cite{potlapalli2023promptir} employ prompt learning with different scenario descriptors for all-in-one restoration, and OneRestore~\cite{guo2024onerestore} further explores composite degradation restoration by improving scene descriptors (see \cref{fig:paradigm}). 
However, inaccurate judgment of scene descriptors affects the application of these solutions.

\subsection{Prompt Learning}

Recently, prompt learning in Natural Language Processing (NLP) \cite{petroni2019language,brown2020language,lester2021power} has gained increasing importance in computer vision \cite{zhou2022learning,ge2023domain}. Pre-trained models can perform various tasks based on prompts containing specific knowledge.
Notably, CoOp~\cite{zhou2022coop} and CoCoOp~\cite{zhou2022cocoop} introduce prompt learning for optimizing CLIP~\cite{radford2021learning}, while SAM~\cite{kirillov2023segment} uses multimodal prompts for unified semantic segmentation. MAE-VQGAN~\cite{bar2022visual} and Painter~\cite{wang2023images} also employ prompts to address multiple visual tasks. PromptIR~\cite{potlapalli2023promptir} and TransWeather~\cite{valanarasu2022transweather} utilize learnable prompts for image restoration across various impaired scenarios, and VPT~\cite{jia2022vpt} introduces visual prompts for this purpose.
Building on prompt learning, OneRestore~\cite{guo2024onerestore} constructs text scene descriptors for adaptive composite degradation restoration. However, in all-in-one restoration and composite degradation restoration, the lack of interpretability in visual prompts may lead to confusion between different degradation tasks, while fixed text prompts struggle to adapt to variations of the same degradation in different scenarios.

%% file: sec/3_method.tex
\begin{figure*}[!t]
    \centering
    \includegraphics[width=1\linewidth]{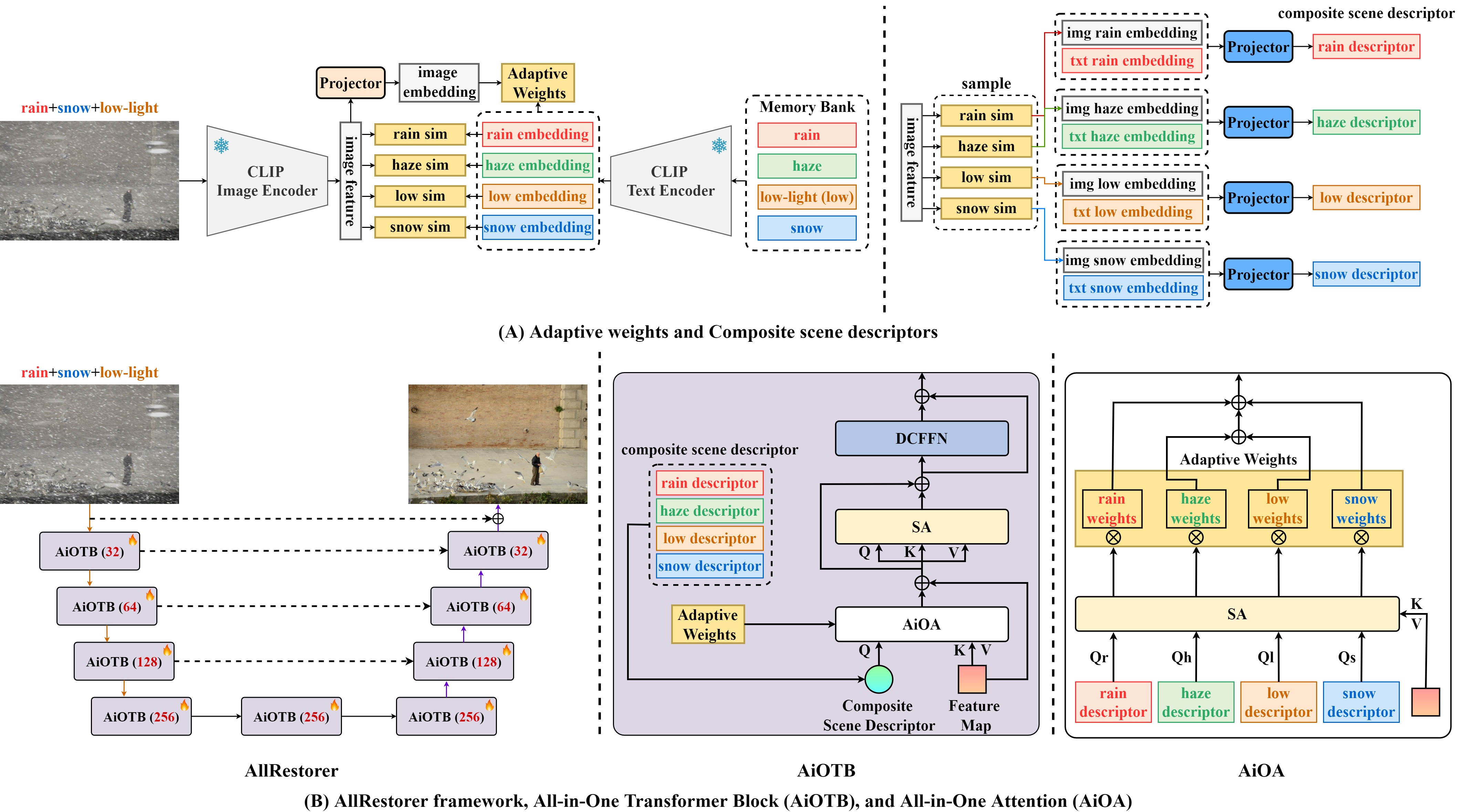}
     \caption{Architecture of AllRestorer. (A) AllRestorer obtains adaptive weights and composite scene descriptors using the fixed CLIP encoder modeling the relationship between each descriptor in the memory bank and the impaired image. (B) In the overall pipeline of AllRestorer, AiOA in AiOTB introduces restoration solutions based on the proportion of each degradation via all composite scene descriptors and adaptive weights. SA in AiOTB adaptively removes all degradations according to the introduced restoration scheme.}
     \label{fig3}
          \vspace{-0.2cm}
   \end{figure*}

\section{The Proposed Method}

In AllRestorer (see \cref{fig3}), we define a memory bank 
$m$ that stores the corresponding texts $t$ for all types of degradation, covering four common physical corrosions: low-light, haze, rain, and snow. To extract these features, we utilize CLIP~\cite{radford2021learning} encoders (fine-tuned on CDD-11) to generate text embeddings for each entry in the memory bank, and image embeddings for the degraded input image.
AllRestorer primarily uses an encoder-decoder U-Net~\cite{ronneberger2015u} framework to capture multi-scale representations. Within this U-Net architecture, we introduce an \textit{All-in-One Transformer Block (AiOTB)} as the core module. Unlike previous all-in-one restoration approaches~\cite{potlapalli2023promptir,guo2024onerestore,chen2024learning}, AiOTB incorporates a novel \textit{All-in-One Attention (AiOA)} mechanism (\cref{3.1}). This mechanism introduces restoration schemes for each degradation by leveraging all scene descriptors as queries.

In AiOTB, \textit{Self-Attention (SA)} is employed to remove the degradations from a given image in the latent space based on the corresponding restoration scheme. Notably, to efficiently characterize the degradation and accommodate its varying manifestations across different scenes, AiOA uses scene descriptors composed of both image and text embeddings (\cref{3.2}). Furthermore, to adapt to varying ratios of different degradations in complex scenes and to ensure that the all-in-one restoration is not negatively affected by unrelated scene descriptors, AiOA integrates an adaptive weight (\cref{3.3}) that adjusts the intensity of restoration for each type of degradation in latent space.
Finally, as in most prior work~\cite{Zamir2021Restormer,huang2023vision,chen2024dual}, AllRestorer incorporates deep convolution in the feed-forward network of AiOTB to effectively restore the local details of the image.

 \subsection{All-in-One Transformer Block}\label{3.1}

To prevent scene descriptors $c$ from being misled by the varying proportions of different degradations in composite degradation scenarios, AllRestorer introduces the novel AiOTB base unit. As illustrated in \cref{fig3}, AiOTB consists of three main components: AiOA, SA, and Deep Convolutional Feed-Forward Network (DCFFN), where the SA function is represented as ${\rm{SA}}(Q,K,V)$.
Unlike attention mechanisms used in previous all-in-one restoration frameworks, AiOA, the core of AiOTB, considers all scene descriptors stored in the memory bank. By modeling long-range dependencies between image features and these descriptors, AiOTB effectively integrates all restoration solutions for the given image:
 \begin{equation}
   \begin{split}
         &{{\rm{AiOA}}(Q_c,K,V)} = {{\lambda}_1}{\rm{SA}}(c_1,K,V)+,\cdots,\\
         &+{{\lambda}_i}{\rm{SA}}(c_i,K,V)+,\cdots,+{{\lambda}_n}{\rm{SA}}(c_n,K,V),
  \end{split}
   \label{eq1}
 \end{equation}
where $\lambda$ represents the adaptive weights defined in \cref{3.3}, which control the restoration intensity for different types of degradation. The variable 
$n$ denotes the total number of degradation types. 
$Q_c$ consists of all scene descriptors used as query matrices, while $K$ and $V$ are the key and value matrices projected from the image features, respectively.
Within AiOTB, SA adaptively selects the appropriate restoration approach for each type of degradation in the latent space, allowing AiOTB to effectively remove all degradations without requiring precise human or model-specific judgments for composite degradations.
For the DCFFN component, we utilize the feed-forward network design from Restormer~\cite{Zamir2021Restormer}.

 \subsection{Composite Scene Descriptor}\label{3.2}

For the AiOA scene descriptor, we develop a composite scene descriptor for AllRestorer. Unlike traditional scene descriptors, our composite version integrates both text and image embeddings. Compared to image descriptors, the inclusion of text embeddings provides a more explicit definition of degradation, helping to reduce confusion across tasks. Meanwhile, unlike text descriptors, the addition of image embeddings allows the composite descriptor to adapt to variations of the same type of degradation across different scenes. Therefore, composite scene descriptors not only avoid ambiguity but also effectively define the degradation characteristics.

Specifically, AllRestorer introduces a fine-tuned CLIP encoder to extract all text embeddings $x_t$ in $m$ and image tokens $e$ for given degraded image $I$, respectively. 
However, for composite degradation, $e$ will encompass various degradations, which interfere with image restoration. 
In AllRestorer we calculate the dot product similarity among all text embeddings and each image token, selecting $k$ image tokens with the highest similarity as the image embeddings $x_m$ of the composite scene descriptor:
\begin{equation}
   \begin{aligned}
         & {e^{1},\cdots,e^{l}} = {{\rm{CLIP}}(I)},\\
         & {x_{t}^{1},\cdots,x_{t}^{n}} = {{{\rm{CLIP}}(t^1)},\cdots,{{\rm{CLIP}}(t^n)}},\\
         & sim^i = softmax(x_{t}^{i}\cdot[{e^{1},\cdots,e^{l}}]),i\in[1,n],\\
         & x_{m}^{i} = sample([{e^{1},\cdots,e^{l}}],sim^i,k),
  \end{aligned}
   \label{eq2}
 \end{equation} 
where $l$ denotes the token number of $e$. 
The function $sample(e, sim, k)$ identifies the $k$ tokens from the image tokens $e$ with the highest dot product similarity $sim$. 
Eventually, the combined image embedding and text embedding will be projected as our composite scene descriptor: 
\begin{equation}
   \begin{aligned}
         {c_i} & = {concat[x_{m}^{i},x_{t}^{i}]}, \quad i\in[1,n],\\
         {c_i} & = {c_i}{W_c},
  \end{aligned}
   \label{eq3}
 \end{equation}
where $W_c$ represents the projection matrix of the composite scene descriptor. 
Depending on the composite scenario descriptor, AllRestorer can define composite degradation more accurately to avoid incorrect restoration.

 \subsection{Adaptive Weights}\label{3.3}

\begin{figure}[t]
    \centering
    \includegraphics[width=1\linewidth]{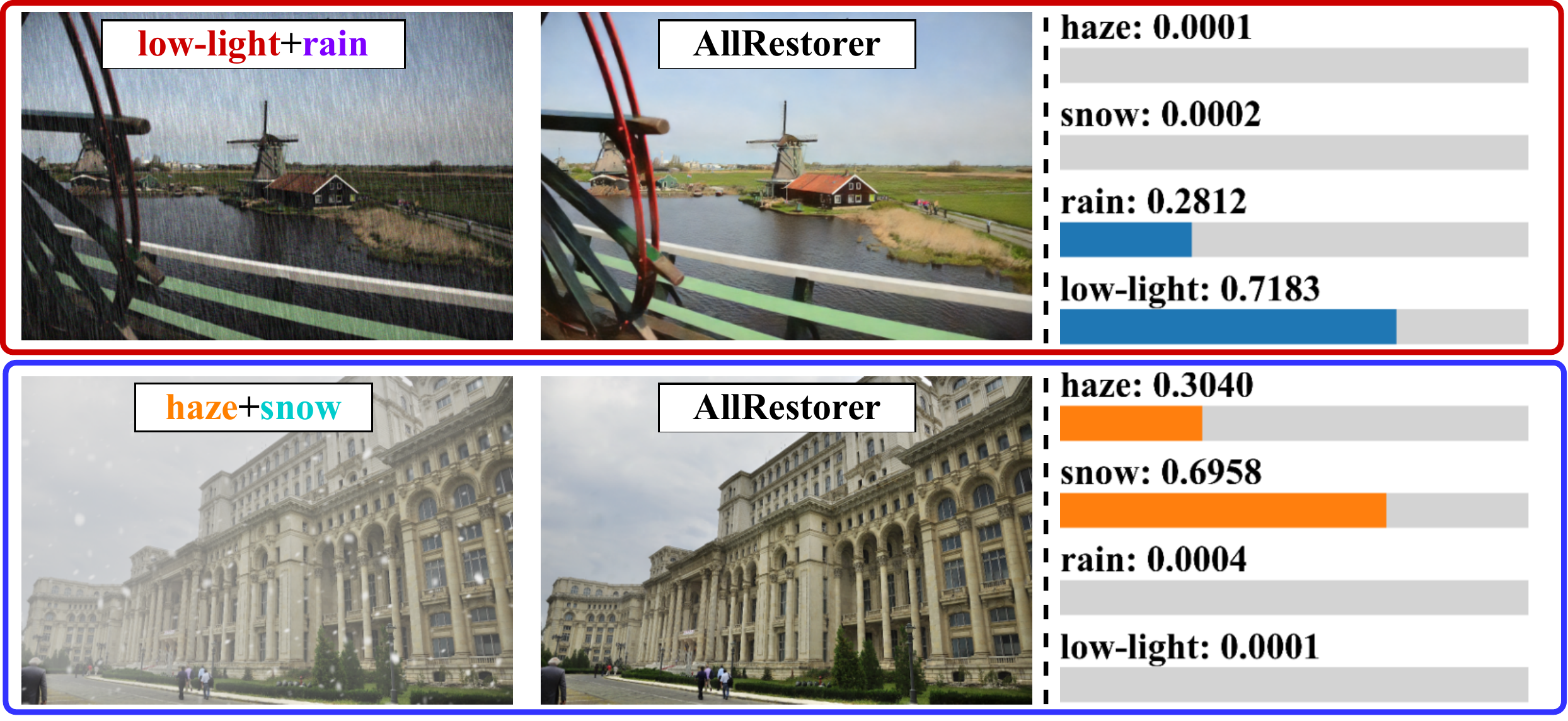}
     \caption{Adaptive weights for each degradation.}
          \vspace{-0.3cm}
 \label{fig:adaptive_weight}
  \end{figure}

To account for the varying proportions of different degradations in composite degradation scenarios and to prevent AiOTB's overall restoration from interfering with single-degradation restoration, AiOA incorporates adaptive weights to regulate the restoration strength for each type of degradation in different contexts (see \cref{fig:adaptive_weight}).

In AllRestorer, the CLIP image projection layer is used to project the image class token $e^1$, thus computing the similarity of the dot product with each text embedding to adaptively represent the proportion of each degradation in the given image. 
The similarity is scaled by the softmax function as an adaptive weight for each degradation: 
\begin{equation}
   \begin{split}
         z &  = {{\rm{CLIP_{proj}}}(e^1)},\\
         {{\lambda}_i} & = softmax(z\cdot{x_{t}^{i}}),i\in[1,n].
  \end{split}
   \label{eq4}
 \end{equation}
Using the similarity between the image projection vectors $z$ and the text embeddings, adaptive weights can adaptively assess the proportion of various degradations in the image, ensuring that AiOTB does not affect the restoration of other degradations while addressing the corresponding one.

%% file: sec/4_experiment.tex
 \section{Experiments}

  \begin{table}
   \centering
    \footnotesize
    \begin{tabular}{c|l|ccc} 
        \toprule
        Types & Method &  PSNR $\uparrow$&  SSIM $\uparrow$ & Params $\downarrow$\\ 
        \midrule
         & MIRNet~\cite{Zamir2020MIRNet} &  25.97 & 0.8474 & 31.79M  \\ 
         & MPRNet~\cite{Zamir2021MPRNet} &  25.47 & 0.8555 & 15.74M  \\
         & MIRNetv2~\cite{Zamir2022MIRNetv2} &  25.37 & 0.8335 & 5.86M  \\ 
         & Restormer~\cite{Zamir2021Restormer} &  28.99 & 0.8646 & 26.13M  \\ 
         \textbf{Uniform}   & DGUNet~\cite{Mou2022DGUNet}& 26.92 & 0.8559 & 17.33M \\
         & NAFNet~\cite{chu2022nafssr} & 24.13 & 0.7964 & 17.11M \\
         & SRUDC~\cite{song2023under} & 27.64 & 0.8600 & 6.80M  \\ 
         & Fourmer~\cite{zhou2023fourmer} & 23.44 & 0.7885 & 0.55M \\ 
         & OKNet~\cite{cui2024omni} & 26.33 & 0.8605 & 4.72M \\ \hline
         & AirNet~\cite{AirNet} & 23.75 & 0.8140 & 8.93M\\ 
         & TransWeather~\cite{valanarasu2022transweather} & 23.13 & 0.7810 & 21.90M\\ 
         \textbf{All-in-One}   & WeatherDiff~\cite{ozdenizci2023} & 22.49 & 0.7985 & 82.96M \\ 
         & PromptIR~\cite{potlapalli2023promptir} & 25.90 & 0.8499 & 38.45M\\ 
         & WGWSNet~\cite{zhu2023Weather} & 26.96 & 0.8626 & 25.76M\\ \hline
         & OneRestore~\cite{guo2024onerestore} & 28.47 & 0.8784 & 5.98M\\
         \textbf{Composite}  & ${\text{OneRestore }^{\star}}$~\cite{guo2024onerestore} & 28.72 & 0.8821 & 5.98M\\
         & AllRestorer (ours) & \textbf{33.72} & \textbf{0.9436} & 12.18M\\
        \bottomrule
    \end{tabular}
    \caption{Quantitative comparison of AllRestorer with baselines on CDD-11, where ${\text{OneRestore}^{\star}}$ denotes OneRestore based on text scene descriptor.}
    \label{tab1}
     \vspace{-0.cm}
\end{table}

 \begin{figure}[t]
    \centering
    \includegraphics[width=1\linewidth]{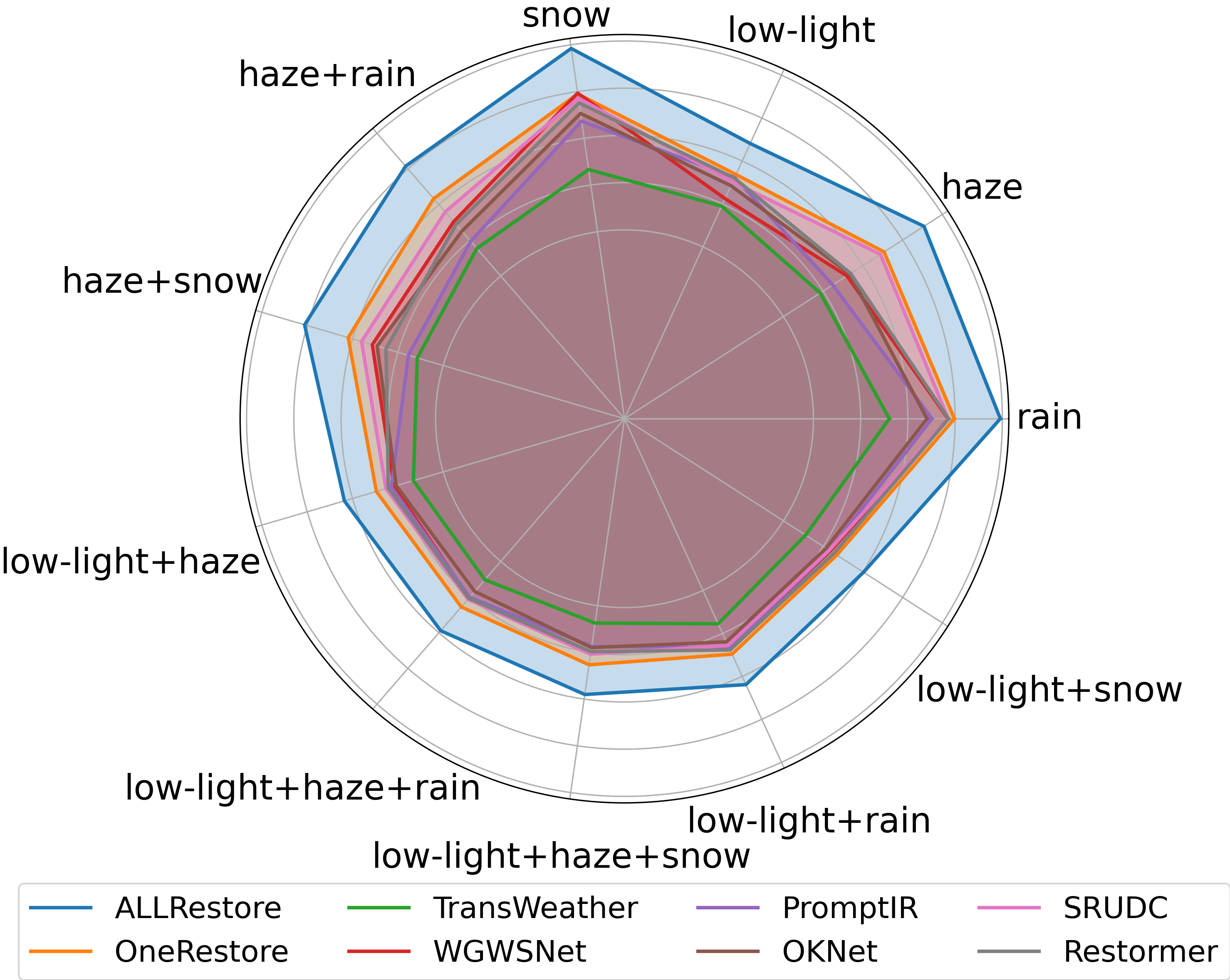}
     \caption{PSNR comparison with uniform baselines and all-in-one baselines for each task in CDD-11.}
 \label{fig4}
      \vspace{-0.2cm}
  \end{figure}

  \begin{figure*}[!t]
    \centering
    \footnotesize
    \def\yem{-2.5pt}
    \def\xwidth{0.99}
    \def\xxxwidth{0.12\textwidth} 
    \setlength{\tabcolsep}{1.3pt}
    
    \begin{tabular}{p{\xxxwidth}p{\xxxwidth}p{\xxxwidth}p{\xxxwidth}p{\xxxwidth}p{\xxxwidth}p{\xxxwidth}p{\xxxwidth}}
        \multicolumn{8}{c}{
            \includegraphics[width=\xwidth\linewidth]{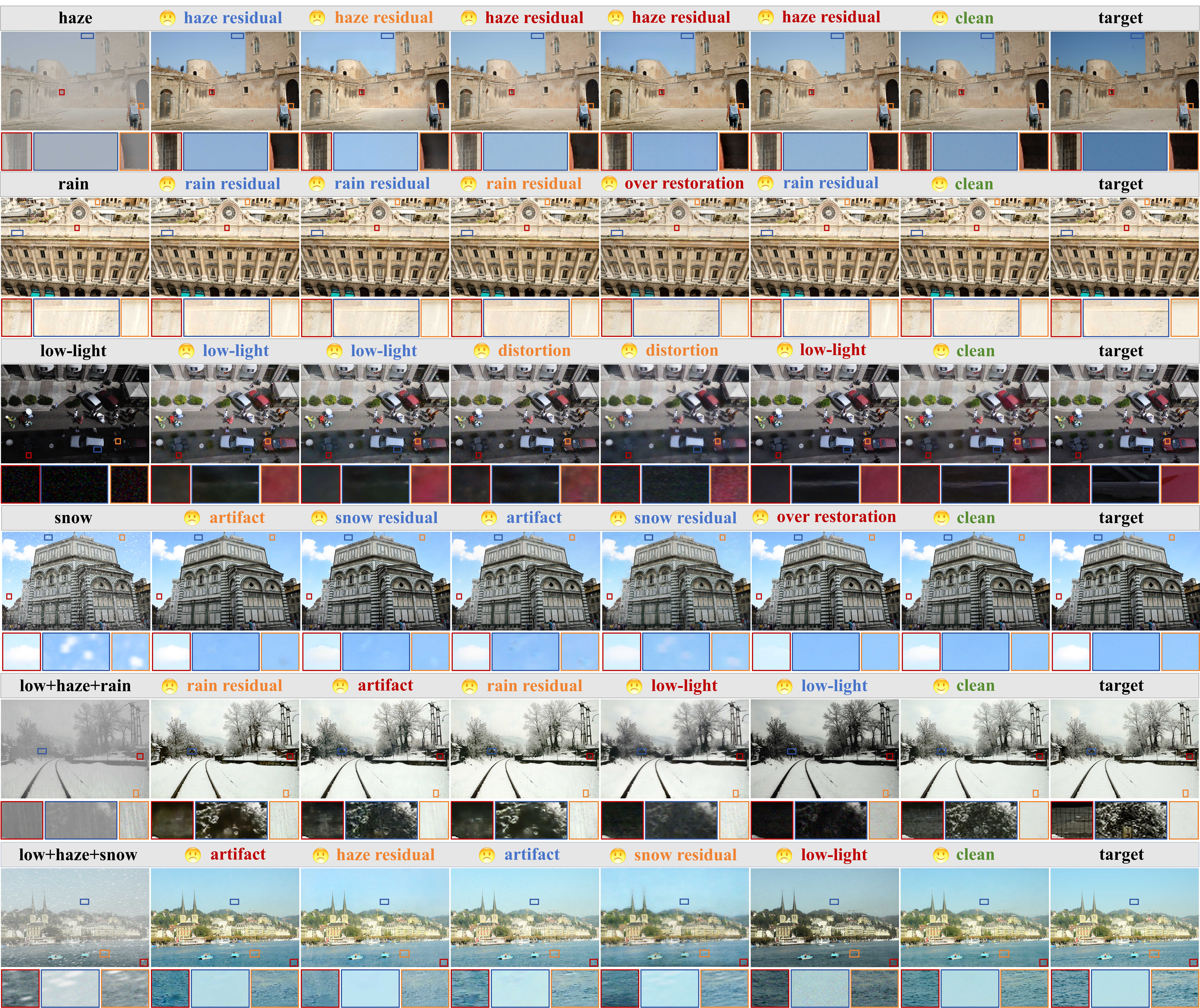}
        } \\ 
        \hspace{24pt} Input & \hspace{6pt} Restormer~\cite{Zamir2021Restormer} & \hspace{11pt} OKNet~\cite{cui2024omni} & \hspace{5pt} PromptIR~\cite{potlapalli2023promptir} & \hspace{-2pt} TransWeather~\cite{valanarasu2022transweather} & \hspace{0pt} OneRestore~\cite{guo2024onerestore} & \hspace{6pt} AllRestorer & \hspace{15pt} Target \\
    \end{tabular}
    \caption{Qualitative comparison of our AllRestorer and baselines on CDD-11~\cite{guo2024onerestore} dataset. Zoom in for a better view.}
    \label{fig:comparison}
     \vspace{-0.1cm}
\end{figure*}

\subsection{Datasets and Evaluation Metrics}

To fairly evaluate the performance of AllRestorer and baselines on all-in-one image restoration and composite degradation restoration tasks, we select the official dataset CDD-11~\cite{guo2024onerestore} for uniform training and test them on synthetic datasets and real datasets. The synthesized data in CDD-11 were utilized to assess all-in-one image restoration and composite degradation restoration. 
Real-world datasets such as LOL-v2~\cite{yang2021sparse}, Snow100k-R~\cite{liu2018desnownet}, NH-HAZE~\cite{Fu_2021_CVPR}, and SPA-Data~\cite{wang2019spatial} were employed to evaluate AllRestorer's performance in real-world all-in-one image restoration. 
Additionally, we gather 200 real composite scene cases from the web (real200) to evaluate AllRestorer’s performance in real-world composite degradation restoration.
For CDD-11 results, we utilize PSNR and SSIM metrics for evaluation. 
In real-world tests, we employ NIQE and PIQE metrics.

 \subsection{Implementation Details}

 We trained AllRestorer for 300 epochs on the PyTorch framework in NVIDIA RTX 3090. 
 The training was optimized using an Adam model with an initial learning rate of 1.25e-4.
 The loss functions include the smoothed $L1$ loss and the perceptual loss based on the VGG model, which is weighted at 0.04. 
 All training images were uniformly set to 256 resolution, and data augmentation was limited to random cropping. 
 Regarding CLIP fine-tuning, we only fine-tuned 40 epochs at 1e-6 learning rate on single degradation data in CDD-11.

 \subsection{Comparison with SOTA Methods}

 We evaluate the performance of AllRestorer, unified image restoration networks (\textit{Uniform}), all-in-one image restoration methods (\textit{All-in-One}), and composite degradation restoration schemes (\textit{Composite}) on CDD-11.

 \begin{figure*}[!t]
\centering
\footnotesize
\def\yem{-2.5pt}
\def\xwidth{0.99}
\def\xxxwidth{0.14\textwidth} 
\setlength{\tabcolsep}{1pt}
\begin{tabular}{p{\xxxwidth}p{\xxxwidth}p{\xxxwidth}p{\xxxwidth}p{\xxxwidth}p{\xxxwidth}p{\xxxwidth}} 
\multicolumn{7}{c}{
\includegraphics[width=\xwidth\linewidth]{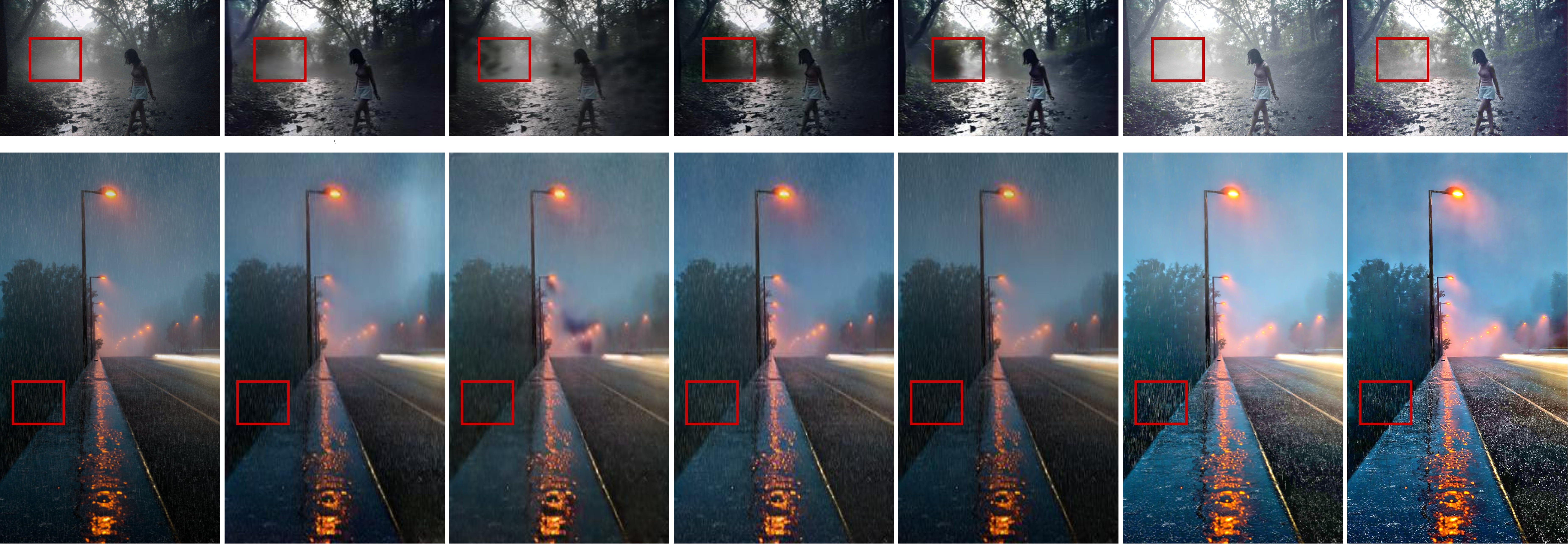} 
} \\ 
\hspace{28pt} Input & \hspace{17pt} OKNet~\cite{cui2024omni} & \hspace{12pt} Fourmer~\cite{zhou2023fourmer} & \hspace{3pt} TransWeather~\cite{valanarasu2022transweather} & \hspace{8pt} PromptIR~\cite{potlapalli2023promptir} & \hspace{3pt} OneRestore~\cite{guo2024onerestore} & \hspace{8pt} AllRestorer
\end{tabular}
\caption{Comparison of composite degradation restoration results between AllRestorer and baselines in the real world.}
\label{fig:comparison_real}
     \vspace{0.4cm}
\end{figure*}

\begin{figure*}[!t]
\centering
\includegraphics[width=1\linewidth]{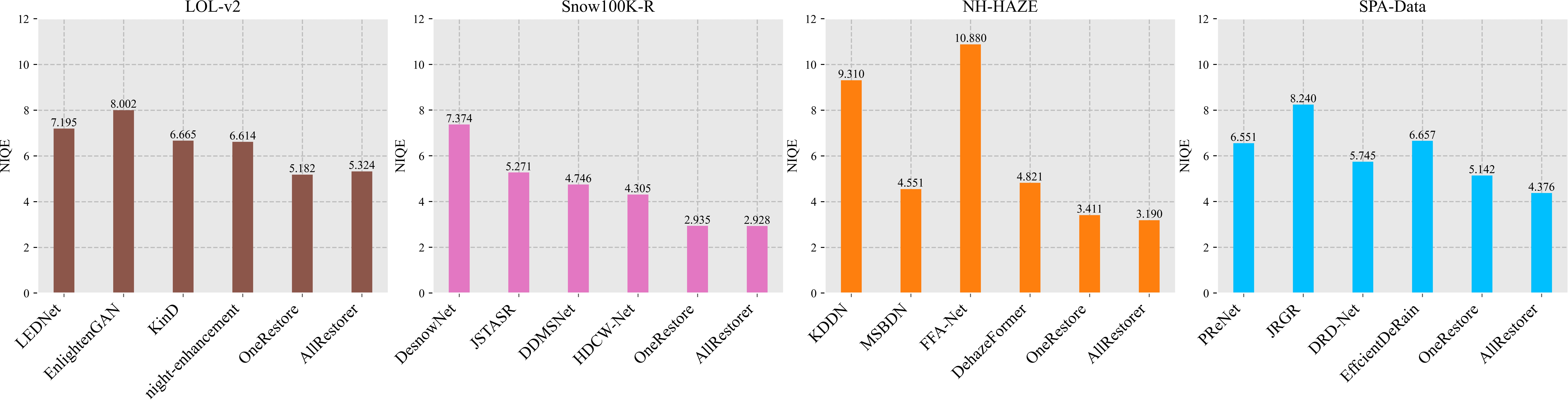}
 \caption{Quantitative comparative results of AllRestorer in real-world all-in-one image restoration.}
      \vspace{-0.2cm}
 \label{fig:comparison_niqe}
\end{figure*}

  \begin{table}
   \centering
    \footnotesize
    \begin{tabular}{l|c|cc} 
        \toprule
        Method &  Venue \& Year & NIQE $\downarrow$&  PIQE $\downarrow$\\ 
        \midrule
         NAFNet~\cite{chu2022nafssr} & ECCV 2022 & 5.561 & 27.0 \\
         SRUDC~\cite{song2023under} & ICCV 2023 & 8.911 & 42.2  \\ 
         Fourmer~\cite{zhou2023fourmer} & ICML 2023 & 5.731 & \textbf{24.9} \\ 
         OKNet~\cite{cui2024omni} & AAAI 2024 & 26.33 & 27.5 \\ 
         AirNet~\cite{AirNet} & CVPR 2022 & 6.283 & 29.5\\ 
         TransWeather~\cite{valanarasu2022transweather} & CVPR 2022 & 5.760 & 27.4\\ 
         PromptIR~\cite{potlapalli2023promptir}& NeurIPS 2023 & 6.103 & 28.2\\ 
         WeatherDiff~\cite{ozdenizci2023} & TPAMI 2023 & 6.201 & 28.8 \\ 
         OneRestore~\cite{guo2024onerestore} & ECCV 2024  & 4.316 & 27.3\\
         AllRestorer (ours) & - & \textbf{4.241} & 26.5\\
        \bottomrule
    \end{tabular}
    \caption{Real world composite degradation restoration results.}
    \label{tab2}
     \vspace{-0.4cm}
\end{table}

\paragraph{Quantitative Evaluation. } In \cref{tab1}, we provide comparison results of Restorer with baselines in CDD-11. 
Quantitative comparison results indicate that AllRestorer, which removes all degradation in the latent space, demonstrates a clear advantage on CDD-11. 
Specifically, AllRestorer achieved 33.72 dB PSNR and 0.9436 SSIM on CDD-11, marking a 5.00 dB PSNR and 0.0615 SSIM improvement over the current state-of-the-art (SoTA) baseline. 
Furthermore, \cref{fig4} illustrates that AllRestorer performs better on each restoration task in CDD-11.

 \paragraph{Qualitative Evaluation.} We present the visual performances of AllRestorer and other methods for all-in-one image restoration and composite degradation restoration in \cref{fig:comparison}. 
 Visual comparisons demonstrate that AllRestorer effectively removes both individual and complex composite degradations. 
 Residual degradations are observed in the results of both unified image restoration baselines and all-in-one restoration baselines. 
 OneRestore occasionally misidentifies the degradation, causing incorrect restoration. 

 \subsection{Real-World Test}

 \begin{figure*}[!t]
    \centering
    \includegraphics[width=1\linewidth]{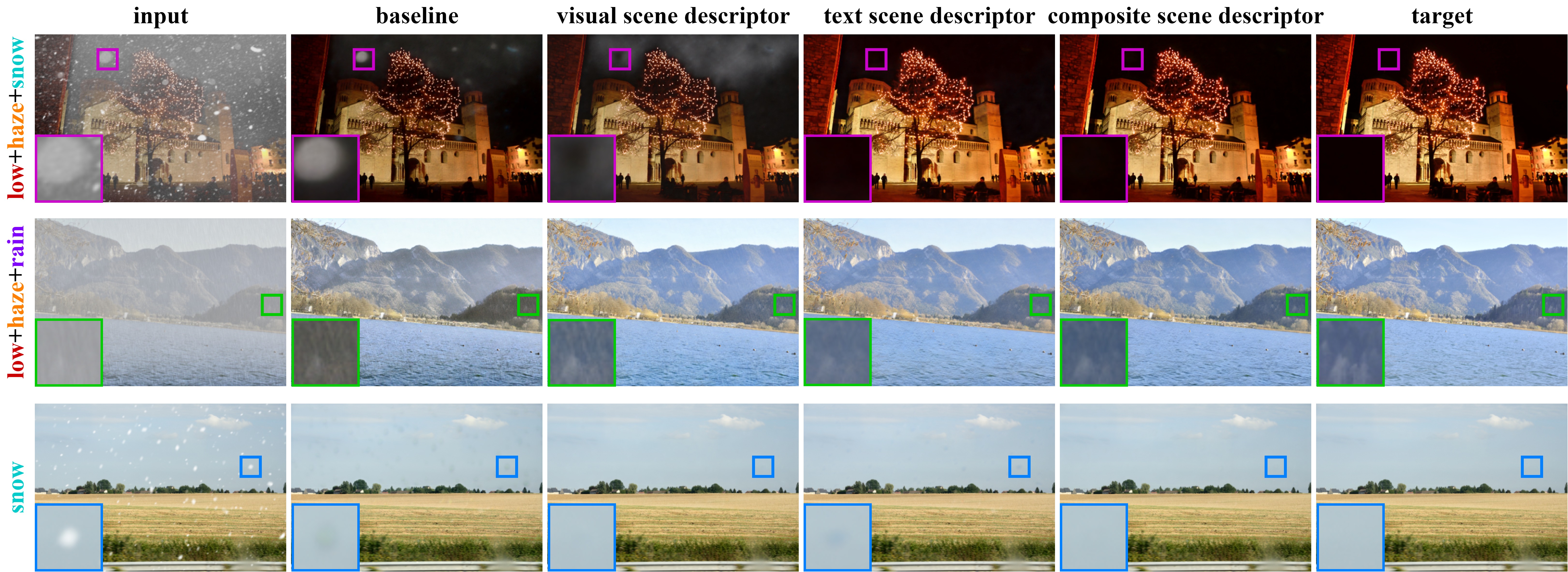}
     \caption{The impact of different scene descriptors on AllRestorer. Zoom in for a better view.}
     \label{fig7}
          \vspace{-0.4cm}
   \end{figure*}

 In this section, we assess AllRestorer's performance in real-world applications. 
 Specifically, we evaluated its restoration capabilities on real-world composite degradation scenarios using 200 real-world cases collected online and tested its all-in-one restoration results under multiple official real-world datasets. 

 \paragraph{Composite Degradation Restoration.} To fairly reflect real-world composite degradation repair performance, we uniformly compared restoration results from AllRestorer and baselines trained on CDD-11 in the real200 examples. 
 \cref{tab2} shows that AllRestorer achieves satisfactory performance in the restoration of composite degraded scenes in the real world. 
 Compared to the baseline method OneRestore, AllRestorer decreases the NIQE by 0.075. 
 In \cref{fig:comparison_real}, we provide a visual comparison of the composite degradation restoration results for different methods in real200 examples. 
 Despite OneRestore enhanced light, rain and haze residuals remain (see \textcolor{red}{red box} in \cref{fig:comparison_real}), and other approaches struggle with real composite degradation scenes. 
 In contrast, AllRestorer achieved more satisfactory results.

 \paragraph{All-in-One Image Restoration. }To evaluate the robustness of AllRestorer in real-world all-in-one restoration, we tested AllRestorer trained on CDD-11 on multiple real benchmarks. 
 For fair comparisons, we also trained the baseline models on the corresponding degraded data in CDD-11. 
 In \cref{fig:comparison_niqe}, we compare AllRestorer with low-light enhancement baselines~\cite{zhou2022lednet,jiang2021enlightengan,zhang2019kindling,sharma2021nighttime}, deraining baselines~\cite{ren2019progressive,deng2020detail,ye2021closing,guo2020efficientderain}, dehazing baselines~\cite{hong2020distilling,MSBDN-DFF,qin2020ffa,song2023vision}, and desnowing baselines~\cite{JSTASRChen,chen2021all,zhang2021deep,liu2018desnownet}, respectively. 
 Compared to baselines, AllRestorer exhibits better restoration in most impaired scenarios. 

 \subsection{Efficiency Comparison}

\begin{table}
   \centering
    \footnotesize
    \begin{tabular}{l|c|cc} 
        \toprule
        Types & Method &  MACs $\downarrow$&  Inference Time $\downarrow$\\ 
        \midrule
         & MPRNet~\cite{Zamir2021MPRNet} &  1393.831G & 75.314ms \\
         & MIRNetv2~\cite{Zamir2022MIRNetv2} &  140.918G & 41.391ms \\ 
         \textbf{Uniform}   & DGUNet~\cite{Mou2022DGUNet}& 200.464G & 39.724ms \\
         & Fourmer~\cite{zhou2023fourmer} & 34.114G & 31.590ms \\ 
         & OKNet~\cite{cui2024omni} & 39.549G & 9.382ms \\ \hline
         & TransWeather~\cite{valanarasu2022transweather} & 6.131G & 12.717ms \\ 
         \textbf{All-in-One}  & PromptIR~\cite{potlapalli2023promptir} & 158.140G & 94.330ms \\ 
         & WGWSNet~\cite{zhu2023Weather} & 1.677G & 69.746ms \\ \hline
         & OneRestore~\cite{guo2024onerestore} & 11.340G & 16.081ms \\
         \textbf{Composite}  & ${\text{OneRestore}^{\star}}$~\cite{guo2024onerestore} & 11.340G & 16.081ms \\
         & ALLRestore & 11.549G & 25.019ms \\
        \bottomrule
    \end{tabular}
    \caption{Efficiency comparison of ALLRestore and baselines under 256$\times$256 resolution in NVIDIA RTX 3090.}
    \label{taba1}
\vspace{-6pt}
\end{table}

In this section, we provide an efficiency comparison of ALLRestore with other methods. 
In \cref{taba1}, we evaluated Multiply-Accumulate Operations (MACs) and Inference Time of different methods on NVIDIA RTX 3090 for 256-resolution images.
Compared to most solutions, ALLRestore has pleasing computational efficiency.

    \begin{table}[!t]
\centering
\scriptsize
\setlength{\tabcolsep}{1.2pt}
\renewcommand{\arraystretch}{1.1}
\begin{tabular}[t]{@{}c@{}}

\scriptsize
\setlength{\tabcolsep}{1.2pt}
\begin{tabular}{ccc|cc}
\toprule
 \scriptsize{AiOA}  &  \scriptsize{SA} &  \scriptsize{DCFFN} &  \scriptsize{PSNR} & \scriptsize{SSIM}  \\
\midrule
  &    & \cmark & 24.81 & 0.8607 \\ \arrayrulecolor{lightgray}\hline
  & \cmark  & \cmark & 27.49 & 0.8710   \\ \arrayrulecolor{lightgray}\hline
 \cmark &   & \cmark & 28.26 & 0.8715  \\ \arrayrulecolor{lightgray}\hline
 \cmark & \cmark & \cmark & \textbf{33.72} & \textbf{0.9436} \\ 
\arrayrulecolor{black}\bottomrule
\multicolumn{5}{c}{{A. Individual components.}}\\
\end{tabular}

\hfill
\scriptsize
\setlength{\tabcolsep}{1.2pt}
\begin{tabular}{ccc|cc}
\toprule
\scriptsize{Visual\textsubscript{$sd$}} & \scriptsize{Text\textsubscript{$sd$}} & \scriptsize{Composite\textsubscript{$sd$}} & \scriptsize{PSNR} & \scriptsize{SSIM}   \\
\midrule
  &   &   & 29.15 & 0.9239  \\ \arrayrulecolor{lightgray}\hline
 \cmark &   &  & 31.38 & 0.9310   \\ \arrayrulecolor{lightgray}\hline
  & \cmark &  & 32.41 & 0.9359 \\ \arrayrulecolor{lightgray}\hline
  &  & \cmark & \textbf{33.72} & \textbf{0.9436}\\
\arrayrulecolor{black}\bottomrule
\multicolumn{5}{c}{{B. Effects of composite scene descriptor.}}\\

\end{tabular} 
\\

\end{tabular}
\caption{Ablation studies for AiOTB and composite scene descriptor, where \textit{sd} indicates scene descriptor}
\label{tab:ablation}
     \vspace{-0.2cm}
\end{table}

 \subsection{Ablation Study}

 \begin{figure}[t]
    \centering
    \includegraphics[width=1\linewidth]{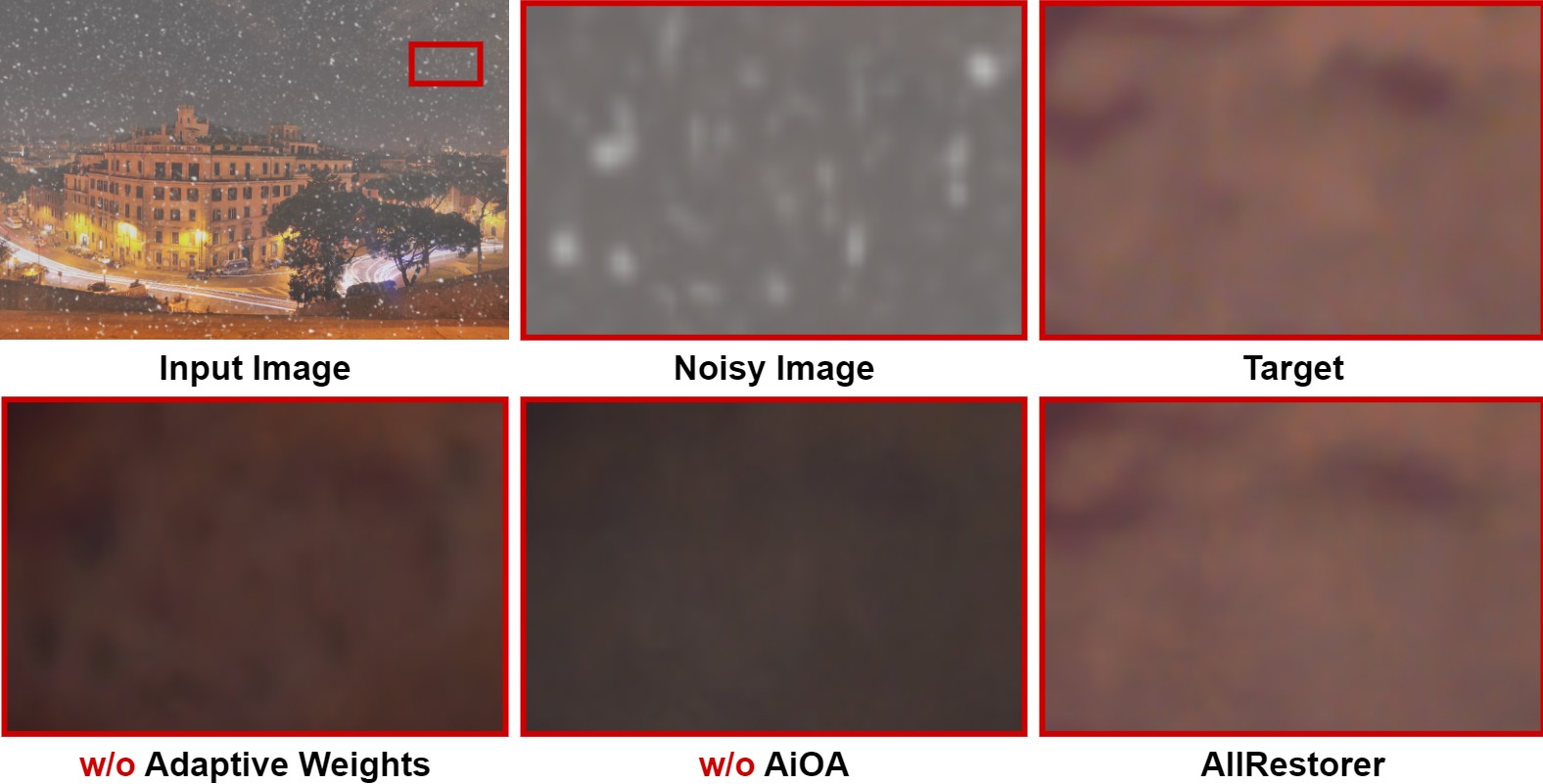}
     \caption{The effect of different components on AllRestorer.}
          \vspace{-0.6cm}
 \label{fig:ablation_component}
  \end{figure}

\paragraph{All-in-One Transformer Block.} We quantitatively evaluate the contribution of each component within AiOTB to AllRestorer's performance on CDD-11.  
\cref{tab:ablation} (A) demonstrates that the baseline using only DCFFN is insufficient for effectively removing degradation even though it can reconstruct the image.
SA improves image restoration but lacks definite guidance. 
While AiOA introduces all restoration solutions for AiOTB, it relies on SA to execute the corresponding restoration for the impaired images. 
Thus, each component in the AiOTB is indispensable (see \cref{fig:ablation_component}).

 \paragraph{Composite Scene Descriptor. }We compare the impact of different scene descriptors on CDD-11 to verify the effectiveness of our composite scene descriptor. \cref{tab:ablation} (B) and \cref{fig7} illustrate this comparison for different scene descriptors. 
 Although baseline and visual scene descriptors can mitigate the effects of a single impairment on image restoration, they are less effective in distinguishing scenes with composite degradations. 
 Text scene descriptors can define different degradations but struggle to adapt to different variations of the same degradation (\textit{e.g., ``snow particle artifacts of some sizes in \cref{fig7}''}). 
 Also, we report the effect of varying image token numbers $k$ on restoration in \cref{tab4}.

 \begin{table}
   \centering
    \footnotesize
    \begin{tabular}{c|cc}
\toprule
 \scriptsize{Setting} &  \scriptsize{PSNR} & \scriptsize{SSIM}  \\
\midrule
  \scriptsize{\textit{w/o} Adaptive weights} & 30.66 & 0.9271 \\ \arrayrulecolor{lightgray}\hline
  \scriptsize{\textit{w/ } Adaptive weights} & \textbf{33.72} & \textbf{0.9436} \\ \arrayrulecolor{black}\bottomrule
 \scriptsize{Composite scene descriptor, $k$=5} & 33.04 & 0.9388   \\ \arrayrulecolor{lightgray}\hline
 \scriptsize{Composite scene descriptor, $k$=10} & \textbf{33.72} & \textbf{0.9436}  \\ \arrayrulecolor{lightgray}\hline
 \scriptsize{Composite scene descriptor, $k$=25} & 33.59 & 0.9431  \\ \arrayrulecolor{lightgray}\hline
 \scriptsize{Composite scene descriptor, $k$=50} & 33.16 & 0.9390  \\  
\arrayrulecolor{black}\bottomrule
\end{tabular}
    \caption{Effect of adaptive weights and the impact of different number sampling tokens $k$ on composite scene descriptor.}
    \label{tab4}
     \vspace{-0.5cm}
\end{table}

 \paragraph{Adaptive Weights.} On CDD-11, we evaluate the contribution of adaptive weights to AllRestorer. 
 As shown in \cref{tab4}, removing adaptive weights notably diminishes AllRestorer’s performance. 
 The absence of a restoration intensity control results in pronounced artifacts in \cref{fig:ablation_component}.
 This indicates that managing restoration intensity for each degradation helps achieve a higher-quality restoration.

%% file: sec/5_conclusion.tex
\section{Conclusion}

In this paper, we propose AllRestorer, a model designed for accurate composite degradation handling and all-in-one image restoration. We develop a novel All-in-One Transformer Block (AiOTB) within AllRestorer, which integrates multiple restoration techniques for degraded images through All-in-One Attention (AiOA). This attention mechanism is guided by composite scene descriptors to effectively remove all types of degradation in the latent space.
AiOA employs adaptive weights to control the restoration intensity for each type of degradation across different scenarios. Leveraging AiOTB's adaptive approach to accurately identify and remove multiple forms of degradation, AllRestorer demonstrates SoTA performance on both synthetic and real datasets.